\documentclass[12pt, anon]{l4dc2024}

\usepackage{todonotes}
\usepackage{wrapfig}
\usepackage{times}
\usepackage{tikz}
\usepackage{mathtools}
\usepackage{url}
\def\arxiv{1}

\usetikzlibrary{automata, positioning}

\title[]{On the Uniqueness of Solution for the Bellman Equation of LTL Objectives}

\definecolor{darkgreen}{RGB}{0,127,0}

\author{%
 \Name{Zetong Xuan} \Email{z.xuan@ufl.edu}\\
 \addr University of Florida, Gainesville, FL 32611, USA
 \AND
 \Name{Alper Kamil Bozkurt} \Email{alper.bozkurt@duke.edu}\\
 \addr Duke University, Durham, NC 27708, USA
 \AND
 \Name{Miroslav Pajic} \Email{miroslav.pajic@duke.edu}\\
 \addr Duke University, Durham, NC 27708, USA
 \AND
 \Name{Yu Wang} \Email{yuwang1@ufl.edu}\\
 \addr University of Florida, Gainesville, FL 32611, USA
}

\begin{document}

\maketitle

\begin{abstract}
Surrogate rewards for linear temporal logic (LTL) objectives are commonly utilized in planning problems for LTL objectives. 
In a widely-adopted surrogate reward approach, two discount factors are used to ensure that the expected return approximates the satisfaction probability of the LTL objective. 
The expected return then can be estimated by methods using the Bellman updates such as reinforcement learning. 
However, the uniqueness of the solution to the Bellman equation with two discount factors has not been explicitly discussed. 
We demonstrate with an example that when one of the discount factors is set to one, as allowed in many previous works, the Bellman equation may have multiple solutions, leading to inaccurate evaluation of the expected return. 
We then propose a condition for the Bellman equation to have the expected return as the unique solution, requiring the solutions for states inside a rejecting bottom strongly connected component (BSCC) to be $0$. 
We prove this condition is sufficient by showing that the solutions for the states with discounting can be separated from those for the states without discounting under this condition.
\end{abstract}


\section{Introduction}
\label{sec:intro}
Modern autonomous systems need to solve planning problems for complex rule-based tasks that
are usually expressible by linear temporal logic (LTL) \citep{pnueli1977temporala}. LTL is a symbolic language that helps fully automate the design process with computer algorithms. 
When the planning environment can be modeled by Markov decision processes (MDPs), the planning problems of finding the optimal policy to maximize the probability of achieving an LTL objective can be solved by model checking techniques \citep{baier2008principles,fainekos2005temporal,kressgazit2009temporallogicbased}. 

However, the utility of model checking is limited when the transition probabilities of the MDP model are unknown. A promising solution, in such scenarios, is to deploy reinforcement learning (RL) \citep{sutton2018reinforcement} to find the optimal policy from sampling.
Early efforts in this direction have been confined to particular subsets of LTL (e.g.~\cite{li2017reinforcement,li2019temporal,cohen2023temporal}), relied restricted
semantics (e.g.~\cite{littman2017environmentindependent}), 
or assumed prior knowledge of the MDP's topology (e.g.~\cite{fu2014probably}) 
--- understanding the presence or absence of transitions between any two given states. 
Model-based RL methods have also been applied by first estimating all the transitions of the MDP and applying model checking with a consideration on the estimation error \citep{brazdil2014verification}. However, the computation complexity can be unnecessarily high since not all transitions are equally relevant \citep{ashok2019paca}.
 
Recent works have used model-free RL for LTL objectives on MDPs with unknown transition probabilities \citep{sadigh2014learning, hasanbeig2019reinforcement, hahn2020faithful, bozkurt2020control}. These approaches are all based on constructing $\omega$-regular automata for the LTL objectives and translating the LTL objective into surrogate rewards within the product of the MDP and the automaton. The surrogate rewards yield the Bellman equations for the satisfaction probability of the LTL objective for a given policy, which can be solved through sampling by RL.

The first approach \citep{sadigh2014learning} employs Rabin automata to transform LTL objectives into Rabin objectives, which are then translated into surrogate rewards, assigning constant positive rewards to certain ``good'' states and negative rewards to ``bad'' states. However, this surrogate reward function is not technically correct, as demonstrated in \citep{hahn2019omegaregular}. The second approach \citep{hasanbeig2019reinforcement} employs limiting deterministic B\"uchi automata to translate LTL objectives into surrogate rewards that assign a constant reward for ``good'' states with a constant discount factor. This approach is also technically flawed, as demonstrated by \citep{hahn2020faithful}. The third method \citep{bozkurt2020control} also utilizes limiting deterministic B\"uchi automata but introduces surrogate rewards featuring a constant reward for ``good'' states and two discount factors that converge to $1$ throughout the training process. 

In more recent works \citep{voloshin2023eventual,shao2023sample,cai2021modular,hasanbeig2023certified}, the surrogate reward with two discount factors from \citep{bozkurt2020control} was used while allowing one discount factor to be equal to $1$. We noticed that in this case, the Bellman equation may have multiple solutions, as that discount factor of $1$ does not provide contraction in many states for the Bellman operator. 
Consequently, the RL algorithm may not converge or may converge to a solution that deviates from the satisfaction probabilities of the LTL objective, leading to non-optimal policies.
To illustrate this, we present a concrete example.
To identify the satisfaction probabilities from the multiple solutions, we propose a sufficient condition that requires the solution of the Bellman equation to be $0$ on all rejecting BSCCs, in which the discount factor is always $1$. 

We show that, under this sufficient condition, the Bellman equation has a unique solution that approximates the satisfaction probabilities for LTL objectives by the following procedure. 
When one of the discount factors equals $1$, we partition the state space into states with discounting and states without discounting based on surrogate reward.
In this case, we first characterize the relationship between all states with discounting and show that their solution is unique since the Bellman operator always has contractions in these states. Then, we show that the whole solution is unique since the solution on states without discounting is uniquely determined by states with discounting.

\section{Preliminaries} \label{sec:prelim}

This section introduces preliminaries on labeled Markov decision processes, linear temporal logic, and probabilistic model
checking.

\subsection{Labeled Markov Decision Processes}\label{sec:mdp} 


We use labeled Markov decision processes (LMDPs) to model planning problems where each decision has a potentially probabilistic outcome. LMDPs augment standard Markov decision processes \citep{baier2008principles} with state labels, enabling assigning properties, such as safety and liveness, to a sequence of states. 

\begin{definition} \label{def:mdp}
A labeled Markov decision process is a tuple $\mathcal{M} = (S, A, P, s_\mathrm{init}, \allowbreak \Lambda, L)$ where
\begin{itemize}
    \setlength{\itemsep}{0pt}
    \item $S$ is a finite set of states and $s_\mathrm{init} \in S$ is the initial state,
    \item $A$ is a finite set of actions 
    where $A(s)$ denotes the set of allowed actions in the state $s\in S$,
    \item $P: S \times A \times S \to [0,1]$ is the transition probability function such that for all $s\in S$, we have 
    \[
        \sum_{s'\in S}P(s,a,s') = \begin{cases}
            1, & a \in A(s) \\
            0, & a \notin A(s)
        \end{cases},
    \]
    \item $\Lambda$ is a finite set of atomic propositions and $L: S \to 2^{\Lambda}$ is a labeling function. 
\end{itemize}
\end{definition}
A path of the LMDP $\mathcal{M}$ is an infinite state sequence $\sigma = s_0 s_1 s_2 \cdots$ such that for all $i \ge 0$, there exists $a_i \in A(s)$ and $s_{i}, s_{i+1} \in S$ with $P(s_i,a_i,s_{i+1}) > 0$. We can construct a corresponding semantic path as $L(\sigma) = L(s_0)L( s_1)\cdots$ by the labeling function $L(s)$. Given a path $\sigma$, the $i$th state is denoted by $\sigma[i] = s_i$. We denote the prefix by $\sigma[{:}i] = s_0 s_1\cdots s_i$ and suffix by $\sigma[i{+}1{:}] = s_{i+1} s_{i+2}\cdots$. 

\subsection{Linear Temporal Logic and Limit-Deterministic B\"uchi Automata} \label{sec:ltl}
In an LMDP $\mathcal{M}$, whether a given semantic path $L(\sigma)$ satisfies a property such as avoiding unsafe states can be expressed using Linear Temporal Logic (LTL).
LTL can specify the change of labels along the path by connecting Boolean 
variables over the labels with two propositional operators, negation $(\neg)$ and conjunction $(\wedge)$, two temporal operators, next $(\bigcirc)$ and until $(\cup)$. 
\begin{definition} \label{def:ltl} 
The LTL formula is defined by the syntax
\begin{align}
\varphi ::= {\rm true}\,|\, \alpha \,|\, \varphi_1 \wedge \varphi_2\,|\neg \varphi \,|\bigcirc \varphi\,|\, \varphi_1 \cup \varphi_2, \alpha \in \Lambda
\end{align}
Satisfaction of an LTL formula $\varphi$ on a path $\sigma$ of an MDP (denoted by $\sigma \models \varphi$) is defined as, 
$\alpha\in \Lambda$ is satisfied on $\sigma$ if $\alpha\in L(\sigma[1])$, 
$\bigcirc \varphi$ is satisfied on $\sigma$ if $\varphi$ is satisfied on $\sigma[1{:}]$, 
$\varphi_1 \cup \varphi_2$ is satisfied on $\sigma$ if there exists $i$ such that $\sigma[i{:}] \models \varphi_2$ and for all $j<i,\,\sigma[j{:}]\models \varphi_1$.
\end{definition}

Other propositional and temporal operators can be derived from previous operators, e.g., (or) $\varphi_1 \vee \varphi_2 \coloneqq \neg(\neg \varphi_1 \wedge \neg \varphi_2)$, (eventually) $\lozenge \varphi \coloneqq {\rm true} \cup \varphi$ and (always) $\square \varphi \coloneqq \neg \lozenge \neg \varphi$. 

We can use Limit-Deterministic B\"uchi Automata (LDBA) to check the satisfaction of an LTL formula on
a path. 
\begin{definition}\label{sec:ldba}
An LDBA is a tuple $\mathcal{A} = (\mathcal{Q},\Sigma,\delta,q_0,B)$ 
where 
$\mathcal{Q}$ is a finite set of automaton states,
$\Sigma$ is a finite alphabet,
$\delta:\mathcal{Q}\times (\Sigma \cup \{\epsilon\})\to 2^\mathcal{Q}$ is a (partial) transition function,
$q_0$ is an initial state,  and $B$ is a set of accepting states, $\delta$ is total except for the $\epsilon$-transitions ($|\delta(q,\alpha)|=1$ for all $q\in \mathcal{Q}, \alpha \in \Sigma$), and there exists a bipartition of $\mathcal{Q}$ to an initial and an accepting component $\mathcal{Q}_{\text{ini}}\cup \mathcal{Q}_{\text{acc}} = \mathcal{Q}$ such that
\begin{itemize}\setlength{\itemsep}{0pt}
    \item there is no transition from $\mathcal{Q}_{\text{acc}}$ to $\mathcal{Q}_{\text{ini}}$, i.e.,
    for any $q\in \mathcal{Q}_{\text{acc}}, v\in \Sigma, \delta(q,v)\subseteq \mathcal{Q}_{\text{acc}}$,
    \item all the accepting states are in $\mathcal{Q}_{\text{acc}}$, i.e., $B\subseteq \mathcal{Q}_{\text{acc}}$,
    \item $\mathcal{Q}_{\text{acc}}$ does not have any outgoing $\epsilon$-transitions, i.e., $\delta(q,\epsilon)=\emptyset$ for any $q\in \mathcal{Q}_{\text{acc}}$.
\end{itemize}
A run is an infinite automaton transition sequence $\rho = (q_0,w_0,q_1), (q_1,w_1,q_2) \cdots$ such that for all $i \ge 0$,  $q_{i+1} \in \delta(q_i, w_i)$.
The run $\rho$ is accepted by the LDBA if it satisfies the B\"uchi condition, i.e., $\mathrm{inf}(\rho) \cap B \ne \emptyset$, where $\mathrm{inf}(\rho)$ denotes the set of automaton states visited by $\rho$ infinitely many times. 
\end{definition}

A path $\sigma=s_0s_1\dots$ of an LMDP $\mathcal{M}$ is considered accepted by an LDBA $\mathcal{A}$ if the semantic path $L(\sigma)$ is the corresponding word $w$ of an accepting run $\rho$ after elimination of $\epsilon$-transitions.


\begin{lemma}\citep[Theorem 1]{sickert2016limitdeterministic}  \label{lem:ltl to buchi}
Given an LTL objective $\varphi$, we can construct an 
LDBA $\mathcal{A}_\varphi$ (with labels $\Sigma=2^{\Lambda}$) 
such that a path $\sigma \models \varphi$
if and only if $\sigma$ is accepted by the LDBA $\mathcal{A}_\varphi$. 
\end{lemma}



\subsection{Product MDP} \label{sec:product mdp}

Planning problems for LTL objectives typically requires a (history-dependent) policy, which determines the current action based on all previous state visits.
\begin{definition} \label{def:policy} 
A policy $\pi$ is a function $\pi:S^+ \to  {A}$ such that $\pi(\sigma[{:}n])\in  {A}(\sigma[n])$, where $S^+$ stands for the set all non-empty finite sequences taken from $S$. A memoryless policy is a policy that only depends on the current state $\pi:S \to  {A}$. Given a LMDP $\mathcal{M} = ( S, A, P, s_0, \Lambda, L)$ and a memoryless policy $\pi$, a Markov chain (MC) induced by policy $\pi$ is a tuple $\mathcal{M_\pi}=(S, P_\pi,s_0,\Lambda,L)$ where $P_\pi(s,s')=P(s,\pi(s),s')$ for all $s,s'\in S$.
\end{definition}

Using the LDBA, we construct a product MDP that augments the MDP
state space along with the state space of the LDBA, such that the state of the product MDP encodes both the physical state and the progression of the LTL objective. In this manner, we ``lift'' the planning problem to the product MDP. Given that the state of the product MDP now encodes all the information necessary for planning, the action can be determined by the current state of the product MDP, resulting in history-independent policies
Formally, the product MDP is defined as follows:
\begin{definition} \label{def:product mdp}
A product MDP $\mathcal{M}^{\times} = ( S^{\times} , A^{\times} , P^{\times},s_0^{\times}, B^{\times})$ of an LMDP $\mathcal{M} = ( S, A, P, s_0, \Lambda, L)$ and an LDBA ${\mathcal{A}}= (\mathcal{Q},\Sigma,\delta,q_0,B)$ is defined by
the set of states $S^\times = S \times \mathcal{Q}$, 
the set of actions ${A}^\times = { {A}} \cup \{\epsilon_q|q\in \mathcal{Q}\}$, 
the transition probability function 
\begin{align}
P^\times(\langle s,q\rangle,a,\langle s',q'\rangle) \notag =
\begin{cases}
P(s,a,s') & q'=\delta(q,L(s)), a \notin {{A}}^\epsilon \\
1             &a=\epsilon_{q'}, q'\in \delta(q,\epsilon), s=s'\\
0             &{\rm otherwise} 
\end{cases}, 
\end{align}
the initial state $s_0^\times=\langle s_0,q_0 \rangle$, and the set of accepting states $B^\times=\{\langle s,q \rangle \in S^\times |q\in B\}$. We say a path $\sigma$ satisfies the B\"uchi condition $\varphi_B$ if $\mathrm{inf}(\sigma)\cap B^\times\ne\emptyset$. Here, $\mathrm{inf}(\sigma)$ denotes the set of states visited infinitely many times on $\sigma$.
\end{definition}

The transitions of the product MDP $\mathcal{M}^\times $ are derived by combining the transitions of the MDP $ \mathcal{M} $ and the LDBA $ \mathcal{A} $. Specifically, the multiple $ \epsilon $-transitions starting from the same states in the LDBA are differentiated by their respective end states $ q $ and are denoted as $ \epsilon_q$. These $ \epsilon $-transitions in the LDBA give rise to corresponding $ \epsilon $-actions 
in the product MDP, each occurring with a probability of $ 1 $. The limit-deterministic nature of LDBAs ensures that the presence of these $\epsilon$-actions within the product MDPs does not prevent the quantitative analysis of the MDPs for planning. In other words, any optimal policy for a product MDP induces an optimal policy for the original MDP, as formally stated below.

\begin{lemma}[\cite{sickert2016limitdeterministic}] \label{lem:memoryless}
For given an LMDP $\mathcal{M}$ and an LTL objective $\varphi$, let $\mathcal{A}_\varphi$ be the LDBA derived from $\varphi$ and let $\mathcal{M}^\times$ be the product MDP constructed from $\mathcal{M}$ and $\mathcal{A}_\varphi$,  with the set of accepting states $B^\times$. Then, a memoryless policy $\pi^\times$ that maximizes the probability of satisfying the B\"uchi condition on $\mathcal{M}^\times$, $P_{\sigma^\times} \big(\sigma^\times \models \square \lozenge B^\times \big)$ where $\sigma^\times {\sim} \mathcal{M}_{\pi^\times}^\times$,
induces a finite memory policy $\pi$ that maximizes the satisfaction probability $P_{\sigma \sim \mathcal{M}_{\pi}} \big(\sigma \models \varphi \big)$ on $\mathcal{M}$.
\end{lemma}

    

\section{Problem Formulation}
In the previous section, we have shown LTL objectives on an LMDP can be converted into a B\"uchi condition on the Product MDP. In this section, we focus on a common surrogate reward used for B\"uchi condition proposed in \citep{bozkurt2020control} and study the uniqueness of solution for the Bellman equation of this surrogate reward, which has not been sufficiently discussed in previous work \citep{voloshin2023eventual,hasanbeig2023certified,shao2023sample}.

For simplicity, we drop $\times$ from the product MDP notation and define the satisfaction probability for the B\"uchi condition as
\begin{align}
    P(s \models \square \lozenge B) := P_{\sigma \sim \mathcal{M}_\pi} \big( \sigma \models \square \lozenge B \mid \exists t: \sigma[t]=s \big). 
\end{align}
When the product MDP model is unknown, the traditional model-based method through graph search \citep{baier2008principles} is not applicable. Alternatively, we may use model-free reinforcement learning with a two-discount-factor surrogate reward proposed by \citep{bozkurt2020control} and widely used in
\citep{voloshin2023eventual,shao2023sample,cai2021modular,hasanbeig2023certified,cai2023safe}.
It 
consists of a reward function $R: S\to\mathbb{R}$ and a state-dependent discount factor function $\Gamma: S\to (0,1]$ with $0<\gamma_B<\gamma\le 1$, 
\begin{align}\label{eqn:surrogate}
R (s):= 
\begin{cases}
1-\gamma_{B} & s\in B \\ 
0 & s\notin B 
\end{cases}
,\quad \Gamma(s):= 
\begin{cases}
\gamma_{B} & s\in B \\ 
\gamma & s\notin B 
\end{cases}.
\end{align}
A positive reward is collected only when an accepting state is visited along the path. Suppose the discount factor $\gamma=1$; then, the satisfaction of B\"uchi condition results in a summation of a geometric series equal to one. The probability of whether a path satisfies the B\"uchi condition is equal to how likely such a geometric series exists along a path.


For this surrogate reward, the $K$-step return ($K\in \mathbb{N}$ or $K=\infty$) of a path from time $t\in \mathbb{N}$ is
\begin{align}
G_{t:K}(\sigma) = \sum_{i=0}^{K} R(\sigma[t+i])\cdot \prod_{j=0}^{i-1} \Gamma(\sigma[t+j]), \quad
G_{t}(\sigma) = \lim_{K\to\infty}G_{t:K}(\sigma).
\end{align}
Accordingly, the value function $V_\pi(s)$ is the expected return conditional on the path starting at $s$ under the policy $\pi$. It approximates
the satisfaction probability 
thus serves as a metric for policy. 
\begin{align}\label{eqn:value}
V_\pi(s) &= \mathbb{E}_\pi [G_t(\sigma)\,|\,\sigma[t] =s]\notag \\
&= \mathbb{E}_\pi[G_t(\sigma)\mid\sigma[t]=s, \sigma \models \square \lozenge B] \cdot P(s\models \square \lozenge B) \nonumber\\
    &+ \mathbb{E}_\pi[G_t(\sigma)\mid\sigma[t]=s, \sigma \nvDash \square \lozenge B] \cdot P(s\nvDash \square \lozenge B)
\end{align}

Given a policy, the value function satisfies the Bellman equation.%
\footnote{We call $V_\pi(s)=R(s) + \gamma \sum_{s'\in S}P_\pi (s,s') V_\pi(s')$ as the Bellman equation and~$V^*_\pi(s)= \max_{a\in A(s)}\{R(s) + \gamma \sum_{s'\in S}P (s, a, s') V^*_\pi(s')\}$ as the Bellman optimality equation.} The Bellman equation is derived from the fact that the value of the current state is equal to the expectation of the current reward plus the discounted value of the next state. For the surrogate reward in the equation~\eqref{eqn:surrogate}, the Bellman equation is given as follows: 
\begin{align} \label{eqn:bellman}
V_\pi(s) 
&=
\begin{cases}
 1-\gamma_{B}+\gamma_{B}\sum_{s'\in S}{P_\pi(s,s')V_\pi(s')} & s\in B \\ 
\gamma\sum_{s'\in S}{P_\pi(s,s')V_\pi(s)} & s\notin B 
\end{cases}.
\end{align}

Previous work \citep{voloshin2023eventual,hasanbeig2023certified,shao2023sample} allows $\gamma=1$. 
However, setting $\gamma=1$ can cause multiple solutions to the Bellman equations, raising concerns about applying model-free RL.
This motivates us to study the following problem. 


\begin{quote}
\textbf{Problem Formulation:} 
For given (product) MDP $\mathcal{M}$ from Definition \ref{def:product mdp} and the surrogate reward from~\eqref{eqn:surrogate}, and a policy $\pi$, find the sufficient conditions under which the Bellman equation from \eqref{eqn:bellman} has a unique solution.
\end{quote}
The following example shows the Bellman equation~\eqref{eqn:bellman} has multiple solutions when $\gamma = 1$~\eqref{eqn:surrogate}. An incorrect solution, different than the expected return from \eqref{eqn:value}, hinders accurate policy evaluation and restricts the application of RL and other optimization techniques.

\begin{wrapfigure}{r}{0.5\textwidth}
    \centering
    \begin{tikzpicture}[>=stealth, node distance=2.5cm, on grid, auto]
        \node[state, initial, initial text={}] (A) {$s_1$};
        \node[state, right=of A] (B) {$s_2$};
        \node[state, right=of B] (C) {$s_3$};
        
        \path[->]
            (A) edge[bend left] node[above] {$\alpha$} (B)
            (A) edge[bend right] node[below] {$\beta$} (C)
            (B) edge[loop above, min distance=10mm] node {1} (B)
            (C) edge[loop above, min distance=10mm] node {1} (C);
    \end{tikzpicture}
    \caption{Example of a three-state Markov decision process. The resulting Bellman equation~\eqref{eqn:Ex1} has multiple solutions when $\gamma = 1$ in the surrogate reward \eqref{eqn:surrogate}, which can mislead to the suboptimal actions.} \label{fig:Ex1}
    \vspace{-30pt}
\end{wrapfigure}
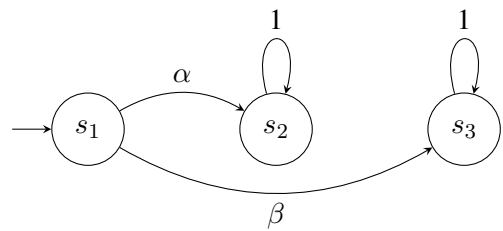

\begin{example} \label{ex:1}
Consider a (product) MDP with three states $S = \{s_1, s_2, s_3\}$ where $s_1$ is the initial state and $B = \{s_2\}$ is the set of accepting states as shown in Figure~\ref{fig:Ex1}. In $s_1$, the action $\alpha$ leads to $s_2$ and the action $\beta$ leads to $s_3$. Since $s_2$ is the only accepting state, $\alpha$ is the optimal action that maximizing the expected return. However, there exists a solution to the corresponding Bellman equation suggesting $\beta$ is the optimal action, as follows: 
\begin{align}\label{eqn:action}
    a^* &\coloneqq \underset{a \in \{\alpha, \beta\}}{\mathrm{argmax}} \{P(s,a,s')V(s')\}  
    = \underset{a \in \{\alpha, \beta\}}{\mathrm{argmax}} \begin{cases}
        V(s_2) & \text{if } a=\alpha, \\
        V(s_3) & \text{if } a=\beta, \\
    \end{cases}
\end{align}
where $V(s_2)$ and $V(s_3)$ can be computed using the Bellman equation~\eqref{eqn:surrogate} as the following: 
\begin{align}\label{eqn:Ex1}
    V(s_2) = 1-\gamma_B + \gamma_B V(s_2), \quad    V(s_3) = V(s_3).
\end{align}
yielding $V(s_2)=1$ and $V(s_3)=c$ where $c\in\mathbb{R}$ is an arbitrary constant.
Suppose $c=2$ is chosen as the solution, then the optimal action will be incorrectly identified as $\beta$ by~\eqref{eqn:action}. 
\end{example} 

\begin{remark}
    The product MDP from Definition~\ref{def:product mdp} is exactly an MDP in the general sense. The surrogate reward~\eqref{eqn:surrogate} and our result based on it work for the general MDPs with B\"uchi objectives.
\end{remark}
\section{Overview of Main Results}
The Bellman equation provides a necessary condition for determining the value function. However, it can have several solutions, with only one being the actual value function (for instance, the Bellman equation for reachability~\citep[P851]{baier2008principles}). It is crucial to identify conditions that eliminate incorrect solutions since solving techniques like model-free RL may struggle to converge or converge to an incorrect solution in the presence of multiple solutions.

In Example~\ref{ex:1}, for $c=0$, the solution for $V(s_3)$ is the value function equal to zero since no reward will be collected on this self-loop based on \eqref{eqn:surrogate}. Generally, the solution should be zero for all states in the rejecting BSCCs, as defined below. 
\begin{definition} \label{def:BSCC}
A bottom strongly connected component (BSCC) of an MC is a strongly connected component without outgoing transitions.
A BSCC is rejecting\footnote{Here we call a state $s\in B$ as an accepting state, a state $s\notin B$ as a rejecting state. Notice that an accepting state must not exist in a rejecting BSCC and a rejecting state may exist in an accepting BSCC.} if all states $s \notin B$. Otherwise, we call it an accepting BSCC. 
\end{definition}

By Definition \ref{def:BSCC}, there will not be any accepting states visited on a path starting from a state in the rejecting BSCCs. Thus, the value function for all states in the rejecting BSCCs equals $0$ based on \eqref{eqn:surrogate}. 
Setting the values for all states within a rejecting BSCC to zero is a sufficient condition for the Bellman equation solution equaling the value function, as stated below.
{\begin{theorem} \label{thm:1}
The Bellman equation~\eqref{eqn:bellman} has the value function as the unique solution, 
if and only if i) the discount factor $\gamma < 1$ or ii) the discount factor $\gamma = 1$ and the solution for any state in a rejecting BSCC is zero. 
\end{theorem}}
\section{Methodology}
We illustrate the proof of Theorem~\ref{thm:1} in this section and provide detailed proofs in \ifx\arxiv\undefined the extended version \citep{xuan2023uniqueness} \else the Appendix\fi.
Specifically, we first prove it for the case of $\gamma<1$ and then move to the case of $\gamma=1$. 
The surrogate reward~\eqref{eqn:surrogate} depends on 
whether a state is an accepting state
or not. 
Thus, we split the state space $S$ by the accepting states $B$ and rejecting states $\neg B:=S\backslash{B}$. The Bellman equation can be rewritten in the following form, 
\begin{align}\label{eqn:bellman vector}
\begin{bmatrix}
V^{B} \\ 
V^{\neg B} \\ 
\end{bmatrix}&=(1-\gamma_{B})
\begin{bmatrix}
\mathbb{I}_{m} \\ 
\mathbb{O}_{n} \\ 
\end{bmatrix}
+\underbrace{\begin{bmatrix}
\gamma_{B}I_{m\times m} &   \\ 
  & \gamma I_{n\times n} \\ 
\end{bmatrix}}_{\Gamma_B}
\underbrace{\begin{bmatrix}
P_{\pi,B\rightarrow B} & P_{\pi,B\rightarrow \neg B} \\ 
P_{\pi,\neg B\rightarrow B} & P_{\pi,\neg B\rightarrow \neg B} \\ 
\end{bmatrix}}_{P_\pi}
\begin{bmatrix}
V^{B} \\ 
V^{\neg B}
\end{bmatrix}, 
\end{align}
where $m=\vert B\vert$, $n=\vert  \neg B \vert$, 
$V^{B}\in\mathbb{R}^{m}$, $V^{\neg B}\in\mathbb{R}^n$ are the vectors listing the value function for all $s\in  {B}$ and $s\in  {\neg B}$, respectively. 
$\mathbb{I}$ and $\mathbb{O}$ are column vectors with all $1$ and $ 0$ elements, respectively. Each of the matrices $P_{\pi,B\rightarrow B}$, $ P_{\pi,B\rightarrow \neg B}$, $P_{\pi,\neg B\rightarrow B}$, $P_{\pi,\neg B\rightarrow \neg B}$ contains the transition probability from a set of states to a set of states, their combination is the transition matrix $P_\pi$ for the induced MC. 
In the following, we assume a fixed policy $\pi$, leading us to omit the $\pi$ subscript from most notation when its implication is clear from the context. 
\subsection{The case \texorpdfstring{$\gamma<1$}{Lg}}
\begin{proposition}\label{prop:1}
If $\gamma<1$ in the surrogate reward~\eqref{eqn:surrogate}, then the Bellman equation~\eqref{eqn:bellman vector} has the value function as the unique solution.
\end{proposition}
As $\gamma<1$, 
the invertibility of
$(I-\Gamma_B P_\pi)$ can be shown by applying Gershgorin circle theorem \citep[Theorem 0]{bell1965gershgorin}. {Any eigenvalue $\lambda$ of $\Gamma_B P_\pi$ satisfies $\vert \lambda\vert <1$ since each row sum of $\Gamma_B P_\pi$ is strictly less than $1$}.
Then, the solution for the Bellman equation~\eqref{eqn:bellman vector} can be uniquely determined as
\begin{align}\label{eqn:matrix}
\begin{bmatrix}
V^{B} \\ 
V^{\neg B} \\ 
\end{bmatrix} 
 &=(1-\gamma_{B})(I_{m+n}-\Gamma_B
P_\pi)^{-1}
\begin{bmatrix}
\mathbb{I}_{m} \\ 
\mathbb{O}_{n} \\ 
\end{bmatrix}.
\end{align} 

\subsection{The case \texorpdfstring{$\gamma=1$}{Lg}}
For $\gamma=1$, the matrix $(I-\Gamma_B P_\pi)$ may not be invertible, causing the Bellman equation~\eqref{eqn:bellman vector} to have multiple solutions. Since the solution may not be the value function here, we use $U^{B}\in \mathbb{R}^m$ and $U^{\neg B}\in \mathbb{R}^n$ to represent a solution on states in $B$ and $\neg B$, respectively. 
In an induced MC, a path starts in an initial state, travels finite steps among the transient states, and eventually enters a BSCC.
If the induced MC has only accepting BSCCs, the connection between all states in $B$ can be captured by a new transition matrix, and the Bellman operator is contractive on the states in $B$. Thus, we can show the solution is unique in all the states in Section~\ref{sec:case1}. 
In the general case where rejecting BSCCs also exists in the MC, 
we introduce a sufficient condition of fixing all solutions within rejecting BSCCs to zero. We demonstrate the uniqueness of the solution under this condition first on $U^{B}$ and then on $U^{\neg B}$ in Section~\ref{sec:case2}.

\subsubsection{When the MC only has accepting BSCCs}\label{sec:case1}
This section focuses on proving that the Bellman equation~\eqref{eqn:bellman vector} has a unique solution when there are no rejecting BSCCs in the MC. 
The result is as follows,
\begin{proposition}\label{prop:2}
        If the MC only has accepting BSCCs and $\gamma=1$ in the surrogate reward~\eqref{eqn:surrogate}, then the Bellman equation~\eqref{eqn:bellman vector} has a unique solution $[{U^B}^T, {U^{\neg B}}^T]^T= \mathbb{I}$. 
\end{proposition}
The intuition behind the proof is to capture the connection between all states $B$ by a new transition matrix $P_\pi^B$, and using $I-\gamma_B P_\pi^B$ invertible to show the solutions $U_B$ 
is unique. Then, we show $U^{\neg B}$ is uniquely determined by
$U^{B}$.

We start with constructing a transition matrix $P_\pi^B$ for the states in $B$ whose $(i,j)$th element, denoted by $(P_\pi^B)_{ij}$, is 
the probability of visiting $j$th state in $B$ without visiting any state in $B$ after leaving the $i$th state in $B$. 
\begin{align}\label{eqn:first return1}
P_\pi^B &:= P_{\pi,B\rightarrow B}+P_{\pi,B\rightarrow \neg B}\sum_{k=0}^\infty {P_{\pi,\neg B\rightarrow \neg B}^{k}}P_{\pi,\neg B\rightarrow B}. 
\end{align}
In \eqref{eqn:first return1}, the matrix element $(P_{\pi,\neg B\to \neg B}^k)_{ij}$ represents the probability of a path leaving the state $i$ and visiting state $j$ after $k$ steps without travelling through any states in $B$. However, the absence of rejecting BSCCs ensures that any path will visit a state in $B$ in finite times with probability $1$. Thus, for any $i, j \in \neg B$, $\lim_{k\to \infty}(P_{\pi,\neg B\to \neg B}^k)_{ij}=0$. 
This limit
implies any eigenvalue $\lambda$ of $P_{\pi,\neg B\to \neg B}$ satisfies $\vert\lambda\vert<1$ and therefore $\sum_{k=0}^\infty {P_{\pi,\neg B\rightarrow \neg B}^{k}}$ can be replaced by $(I - P_{\pi,\neg B\rightarrow \neg B})^{-1}$ in~\eqref{eqn:first return1}, 
\begin{align}\label{eqn:first return}
P_\pi^B =P_{\pi,B\rightarrow B}+P_{\pi,B\rightarrow \neg B} (I - P_{\pi,\neg B\rightarrow \neg B})^{-1}P_{\pi,\neg B\rightarrow B}.
\end{align} 
Since all the elements on the right-hand side are greater or equal to zero, for any $i, j \in B$, $(P_\pi^B)_{ij}\ge 0$. Since there are only accepting BSCCs in the MC, given a path starting from an arbitrary state in $B$, the path will visit an accepting state in finite steps with probability one, ensuring that for all $i\in B$, $\sum_{j\in S} (P_\pi^B)_{ij}=1$. Thus, $P_\pi^B$ is a probability matrix that can be used to describe the behaviour of an MC with the state space as $B$ only. 

\begin{remark}
For a given MC with only accepting BSCCs $\mathcal{M}_\pi=(S, P_\pi,\allowbreak s_0,\Lambda, L)$, we can construct an MC consisting of the accepting states $\mathcal{M}_\pi^B \coloneqq (B, P_\pi^B,\mu,\Lambda, L)$. This new MC, referred to as the accepting MC, has the state space defined as the set of accepting states $B$. The transition probabilities $P_\pi^B$ (from \eqref{eqn:first return}) are the transition probabilities between the accepting states in $\mathcal{M}_\pi$. The initial distribution $\mu$ is a distribution on $B$ and determined by $s_0$ as follows:
\begin{align}
{\rm if}\,  s_0\in B,\quad   \lambda(s) =  
\begin{cases}
1 & s= s_0 \\ 
0 & s\ne s_0 
\end{cases},\quad
{\rm if}\,  s_0\notin B, \quad 
    \mu(s) = (P_{init})_{s_0\, s}
\end{align}
where $P_{init} := (I-P_{\pi,\neg B\to \neg B})^{-1}P_{\pi,\neg B\to B}$ is a matrix. Each element $(P_{init})_{ij}$ represents the probability of a path leaving the state $i\in {\neg B}$ and visiting state $j\in B$ without visiting any state in $B$ between the leave and visit.
Since the absence of rejecting BSCC, we can construct an accepting MC, and every state will have a reward of $1-\gamma_B$ and a discount factor of $\gamma_B$. 
\end{remark}
 
\begin{lemma}\label{lem:1}
    Suppose there is no rejecting BSCC, for $\gamma =1$ in~\eqref{eqn:surrogate}, the Bellman equation~\eqref{eqn:bellman vector} is equivalent to the following form, 
\begin{align}\label{eqn:bellman case1}
\begin{bmatrix}
U^{B} \\ 
U^{\neg B} \\ 
\end{bmatrix} &=(1-\gamma_{B})
\begin{bmatrix}
\mathbb{I}_{m} \\ 
\mathbb{O}_{n} \\ 
\end{bmatrix} +\begin{bmatrix}
\gamma_{B}I_{m\times m} &   \\ 
  & I_{n\times n} \\ 
\end{bmatrix}\begin{bmatrix}
P_\pi^{B} &   \\ 
P_{\pi,\neg B\rightarrow B} & P_{\pi,\neg B\rightarrow \neg B} \\ 
\end{bmatrix}
\begin{bmatrix}
U^{B} \\ 
U^{\neg B} \\ 
\end{bmatrix}. 
\end{align}
\end{lemma}

The equation~\eqref{eqn:bellman case1} implies that the solution $U^{B}$ does not rely on 
the rejecting states $\neg B$. Subsequently, we leverage the fact that $U^{\neg B}$ is uniquely determined by  $U^{B}$
to establish the uniqueness of the overall solution $V$.

Proposition~\ref{prop:2} shows the solutions for the states inside an accepting BSCC have to be $1$. 
All states outside the BSCC cannot be reached from a state inside the BSCC, thus the solution for states outside this BSCC is not involved in the solution for states inside the BSCC. 
By Lemma~\ref{lem:1}, the Bellman equation for an accepting BSCC can be rewritten into the form of~\eqref{eqn:bellman case1} where $U_B$ and $U_{\neg B}$ stands for the solution for accepting states and rejecting states inside this BSCC, and vector $\mathbb{I}$ is the unique solution.

\subsubsection{When accepting and rejecting BSCC both exist in the MC}\label{sec:case2}



Having established the uniqueness of solutions in the case of accepting BSCCs, we now shift our focus to the general case involving rejecting BSCCs. 
We state in Proposition~\ref{prop:2} that the solutions for the states in the accepting BSCCs are unique and equal to $\mathbb{I}$.
We now demonstrate that setting the solutions for the states in rejecting BSCCs to $\mathbb{O}$ ensures the uniqueness and correctness of the solutions for all states.
We partition the state space further into $\{B_A, B_T, \neg B_A, \neg B_R, \neg B_T\}$, where $B_A$ is the set of accepting states in the BSCCs, $B_T:=B\backslash B_A$ is the set of transient accepting states. $\neg B_A$ is the set of rejecting states in the accepting BSCCs, $\neg B_R$ is the set of rejecting states in the rejecting BSCCs, and $\neg B_T:=\neg B\backslash (\neg B_A \cup \neg B_R)$ is set of transient rejecting states. 
We rewrite the Bellman equation~\eqref{eqn:bellman vector} in the following form, 
\begin{align}\label{eqn:Bellman Rej}
&\begin{bmatrix}
U^{B_T} \\ U^{B_A} \\ U^{\neg B_T} \\ U^{\neg B_A} \\ U^{\neg B_R} \\ 
\end{bmatrix} =(1-\gamma_{B})
\begin{bmatrix}
\mathbb{I}_{m} \\ 
\mathbb{O}_{n} \\ 
\end{bmatrix} +\begin{bmatrix}
\gamma_{B}I_{m\times m} &   \\ 
  & I_{n\times n} \\ 
\end{bmatrix} \notag \\ &\begin{bmatrix}
P_{\pi,B_T\rightarrow B_T} & P_{\pi,B_T\rightarrow B_A} & P_{\pi,B_T\rightarrow \neg B_T} & P_{\pi,B_T\rightarrow \neg B_A} & P_{\pi,B_T\rightarrow \neg B_R} \\ 
 & P_{\pi,B_A\rightarrow B_A} & & P_{\pi,B_A\rightarrow \neg B_A}  & P_{\pi,B_A\rightarrow \neg B_R} \\ 
P_{\pi,\neg B_T\rightarrow B_T} & P_{\pi,\neg B_T\rightarrow B_A} & P_{\pi,\neg B_T\rightarrow \neg B_T} & P_{\pi,\neg B_T\rightarrow \neg B_A} & P_{\pi,\neg B_T\rightarrow \neg B_R} \\ 
  & P_{\pi,\neg B_A\rightarrow B_A} &  & P_{\pi,\neg B_A\rightarrow \neg B_A}  & P_{\pi,\neg B_A\rightarrow \neg B_R}\\ 
    & & &  & P_{\pi,\neg B_R\rightarrow \neg B_R}  \\ 
\end{bmatrix}
\begin{bmatrix}
U^{B_T} \\ U^{B_A} \\ U^{\neg B_T} \\ U^{\neg B_A} \\ U^{\neg B_R} \\ 
\end{bmatrix}. 
\end{align}  
The solution for states inside BSCCs has been fixed as $[{U^{B_A}}^T,{U^{\neg B_A}}^T]^T=\mathbb{I}$ and $U^{\neg B_R}=\mathbb{O}$. 
The solution $U^{B_T}$ and $U^{\neg B_T}$ for transient states remain to be shown as unique. We rewrite the Bellman equation~\eqref{eqn:Bellman Rej} into the following form \eqref{eqn:Bellman tran} where $U^{B_T}$ and $U^{\neg B_T}$ are the only variables,
\begin{align}\label{eqn:Bellman tran}
\begin{bmatrix}
U^{B_T} \\ U^{\neg B_T}\\ 
\end{bmatrix} = 
\begin{bmatrix}
\gamma_{B}I_{m_1\times m_1}  &   \\ 
  & I_{n_1\times n_1} \\ 
\end{bmatrix}  \begin{bmatrix}
P_{\pi,B_T\rightarrow B_T} & P_{\pi,B_T\rightarrow \neg B_T}\\ 
P_{\pi,\neg B_T\rightarrow B_T} & P_{\pi,\neg B_T\rightarrow \neg B_T}  \\ 
\end{bmatrix}
\begin{bmatrix}
U^{B_T} \\ U^{\neg B_T}\\ 
\end{bmatrix} +\begin{bmatrix}
B_1 \\ B_2\\ 
\end{bmatrix}
\end{align}   
here $m_1=\vert U^{B_T}\vert$, $n_1=\vert U^{\neg B_T}\vert$ and 
\begin{align}
\begin{bmatrix}
B_1 \\ B_2\\ 
\end{bmatrix} &= 
(1-\gamma_{B})
\begin{bmatrix}
\mathbb{I}_{m_1} \\ 
\mathbb{O}_{n_1} \\ 
\end{bmatrix}  +\begin{bmatrix}
\gamma_{B}I_{m_1\times m_1}  &   \\ 
  & I_{n_1\times n_1} \\ 
\end{bmatrix}\begin{bmatrix}
P_{\pi,B_T\rightarrow B_A} & P_{\pi,B_T\rightarrow \neg B_A}\\ 
P_{\pi,\neg B_T\rightarrow B_A} & P_{\pi,\neg B_T\rightarrow \neg B_A}  \\ 
\end{bmatrix}
\begin{bmatrix}
\mathbb{I}_{m_1} \\ \mathbb{I}_{n_1}\\ 
\end{bmatrix}\notag. 
\end{align}
\vspace{-10pt}
\renewcommand{\proofname}{Proof}
\begin{lemma}\label{lem: unique general}
The equation~\eqref{eqn:Bellman tran} has a unique solution. 
\end{lemma}
We demonstrate $U^{B_T}$ does not rely on 
states in $\neg B_T$ and $U^{\neg B_T}$ 
is uniquely determined by
$U^{B_T}$. Then the uniqueness of $U^{B_T}$ can be shown first, consequently, uniqueness of $U^{\neg B_T}$ can be shown. 

\renewcommand{\proofname}{Proof for Theorem~\ref{thm:1}}
\begin{proof} 
For the case $\gamma=1$, we have shown that the equation~\eqref{eqn:Bellman tran} with surrogate reward~\eqref{eqn:surrogate} has a unique solution in Lemma~\ref{lem: unique general}. In order to complete the proof for theorem~\ref{thm:1}, what remains to be shown is the unique solution of the equation~\eqref{eqn:Bellman tran} is equal to the value function~\eqref{eqn:value}. 

The solution to the equation~\eqref{eqn:Bellman tran} is unique. 
For all $s\in \neg B_R$, the value function $V(s)=0$, then the value function is the unique solution for equation~\eqref{eqn:Bellman tran}.
Under the condition that the solution for all rejecting BSCCs is zero, the Bellman equation~\eqref{eqn:bellman vector} is equivalent to the equation~\eqref{eqn:Bellman tran}. 
We can say theorem~\ref{thm:1} is true.
\end{proof}

\vspace{-20pt}
\section{Conclusion}
This work uncovers a challenge when using surrogate rewards with two discount factors for LTL objectives, which has been unfortunately overlooked by many previous works. Specifically, we show setting one of the discount factors to one can cause the Bellman equation to have multiple solutions, hindering the derivation of the value function.
We discuss the uniqueness of the solution for the Bellman Equation with two discount factors and propose a condition to identify the value function from the multiple solutions. 
Our findings have implications for ensuring convergence to optimal policies in RL for LTL objectives.

\bibliography{References}

\ifx\arxiv\undefined
\else
\newpage
\section*{Appendix}
We provide the following proofs in this appendix.

\subsection*{Proof of Proposition~\ref{prop:1}}\label{proof:1}
The solution of the Bellman equation~\eqref{eqn:bellman vector} can be determined uniquely by matrix operation~\eqref{eqn:matrix} if $(I-\Gamma_B P_\pi)$ is invertible.
The invertibility is shown using the Gershgorin circle theorem \citep[Theorem 0]{bell1965gershgorin}, which claims the following. For a square matrix $A$, 
define the radius as $r_i:=\sum_{j\ne i}{\vert A_{ij}\vert}$. Then, each eigenvalue of $A$ is in at least one of the Gershgorin disks $\mathcal{D}(A_{ii},r_i):=\{z:\vert z-A_{ii}\vert\le r_i\}$.

For the matrix $\Gamma_B P_\pi$, at its $i$-th row, we have the center of the disk as $(\Gamma_B P_\pi)_{ii}=(\Gamma_B)_{ii}{(P_\pi)}_{ii}$, and the radius as $r_i=\sum_{j\ne i}{\vert (\Gamma_B P_\pi)_{ij}\vert} ={(\Gamma_B)}_{ii}(1-{(P_\pi)}_{ii})$. We can upper bound the disk as
\begin{align}
\mathcal{D}((\Gamma_B P_\pi)_{ii},r_i) &= \{z:\vert z-(\Gamma_B P_\pi)_{ii}\vert\le r_i\}\notag\\
&\subseteq\{z: \vert z\vert \le (\Gamma_B P_\pi)_{ii}+r_i\} \notag\\
&\subseteq\{z: \vert z\vert \le \gamma\}.
\end{align}
Since all Gershgorin disks share the same upper bound, the union of all disks is also bounded by 
\begin{align}
    \bigcup_{i\in S}{\mathcal{D}((\Gamma_B P_\pi)_{ii},r_i)}\subseteq\{z: \vert z\vert \le \gamma\}.
\end{align}
The inequality $\gamma<1$ ensures that any eigenvalue $\lambda$ of $\Gamma_B P$ satisfies $\vert \lambda\vert <1$. 
Thus $(I-\Gamma_B P_\pi)$ is invertible and the solution can be uniquely determined by \eqref{eqn:matrix}. 
The value function satisfies the Bellman equation~\eqref{eqn:bellman vector}, which has a unique solution, which is the value function. Thus, the proposition holds.
\subsection*{Proof of Lemma~\ref{lem:1}}\label{proof:2}
We prove this lemma by showing showing the equivalence between $P_\pi^B U^{B}$ and $P_{\pi,B\to B}U^{B} + P_{\pi,B\to\neg B}U^{\neg B}$. 
From the Bellman equation~\eqref{eqn:bellman vector}, we have $U^{\neg B}=P_{\pi,\neg B\to B} U^{B} + P_{\pi,\neg B\to \neg B}U^{\neg B}$
\begin{align}
&P_{\pi,B\to B}U^{B} + P_{\pi,B\to\neg B}U^{\neg B}  \notag \\
&\overset{\textcircled{a}}{=}(P_{\pi,B\to B}  + P_{\pi,B\to\neg B} P_{\pi,\neg B\to B}) U^{B} + P_{\pi,B\to\neg B} P_{\pi,\neg B\to \neg B}U^{\neg B}\notag\\
&\overset{\textcircled{b}}{=}(P_{\pi,B\to B}  + P_{\pi,B\to\neg B} P_{\pi,\neg B\to B}  + P_{\pi,B\to\neg B}P_{\pi,\neg B\to \neg B} P_{\pi,\neg B\to B}) U^{B} + P_{\pi,B\to\neg B} P_{\pi,\neg B\to \neg B}^2U^{\neg B}\notag\\
&\vdots\notag\\
&\overset{\textcircled{c}}{=}\lim_{K\to \infty}\big{(}(P_{\pi,B\to B}  + P_{\pi,B\to\neg B} \sum_{k=0}^K{P_{\pi,\neg B\to \neg B}^k} P_{\pi,\neg B\to B} )U^{B} + P_{\pi,B\to\neg B} P_{\pi,\neg B\to \neg B}^{K+1}U^{\neg B}\big{)} \notag \\
&\overset{\textcircled{d}}{=}P_{\pi,B}U^{B} .
\end{align}
where the equality $\textcircled{a}$ holds as hold as we replace $U^{\neg B}$ in the last term $P_{\pi,B\to\neg B}U^{\neg B}$ by $P_{\pi,\neg B\to B} U^{B} + P_{\pi,\neg B\to \neg B}U^{\neg B}$. 
Similarily, the equalities $\textcircled{b}$ and $\textcircled{c}$ hold as we keep replacing $U^{\neg B}$ in the last term by $P_{\pi,\neg B\to B} U^{B} + P_{\pi,\neg B\to \neg B}U^{\neg B}$. 
The equality $\textcircled{d}$ holds by the definition of $P_\pi^B$.

\subsection*{Proof of Proposition~\ref{prop:2}}\label{proof:4}
From equation~\eqref{eqn:bellman case1}, we obtain the expression for the sub-MC with only accepting states, 
\begin{equation}\label{eqn:sub-MC}
U^{B} =(1-\gamma_{B})\mathbb{I} +\gamma_{B}P_\pi^B U^{B}.
\end{equation}
Given that all the eigenvalues of $P_\pi^B$ are within the unit disk and $\gamma_B<1$, the matrix $(I-\gamma_B P_\pi^B)$ is invertible. $U^{B}$ is uniquely determined by
\begin{align}\label{eqn:VB}
    U^{B} = (1-\gamma_{B})(I-\gamma_B P_\pi^B)^{-1} \mathbb{I}.
\end{align}
Moving to the set of rejecting states $U^{\neg B}$, 
from equation~\eqref{eqn:bellman case1} we have, 
\begin{align}\label{eqn:neg B}
U^{\neg B} &= P_{\pi,\neg B\rightarrow B}U^{B}+P_{\pi,\neg B\rightarrow \neg B}U^{\neg B} \notag \\
&=(I-P_{\pi,\neg B\rightarrow \neg B})^{-1}P_{\pi,\neg B\rightarrow B}U^{B}.
\end{align}
Given the uniqueness of  $U^{B}$, and the invertibility of $(I-P_{\pi,\neg B\rightarrow \neg B})$, we conclude that $U^{\neg B}$ is also unique. 

Let $U^{B}=\mathbb{I}_m$ and $U^{\neg B}=\mathbb{I}_n$, the Bellman equation~\eqref{eqn:bellman vector} holds as the summation of each row of a probability matrix $P_B$ is always one, 
\begin{align} 
\begin{bmatrix}
\mathbb{I}_m \\ 
\mathbb{I}_n \\ 
\end{bmatrix} &=(1-\gamma_{B})
\begin{bmatrix}
\mathbb{I}_{m} \\ 
\mathbb{O}_{n} \\ 
\end{bmatrix} +\begin{bmatrix}
\gamma_{B}I_{m\times m} &   \\ 
  & I_{n\times n} \\ 
\end{bmatrix}
P_\pi
\begin{bmatrix}
\mathbb{I}_m \\ 
\mathbb{I}_n \\ 
\end{bmatrix}. 
\end{align}
Therefore, in the absence of rejecting BSCCs, the unique solution to the Bellman equation~\eqref{eqn:bellman vector} is $\mathbb{I}$.

\subsection*{Proof of Lemma~\ref{lem: unique general}}\label{proof:3}
First we show $U^{B_T}$ is unique since its calculation can exclude $U^{\neg B}$. 
By equation~\eqref{eqn:Bellman tran}, 
\begin{align}\label{eqn:replace2}
     U^{\neg B_T}=P_{\pi,\neg B_T\to B_T}U^{B_T}+P_{\pi,\neg B_T\to \neg B_T}U^{\neg B_T} +
     \begin{bmatrix}
    P_{\pi,\neg B_T\to B_A} & P_{\pi,\neg B_T\to\neg B_A} 
    \end{bmatrix}
\mathbb{I} 
\end{align}
The following equalities hold as we keep expending $U^{\neg B_T}$, 
\begin{align}
    &P_{\pi,B_T\to B_T}U^{B_T}+P_{\pi,B_T\to\neg B_T}U^{\neg B_T}  \\
    \overset{\textcircled{a}}{=}&(P_{\pi,B_T\to B_T}+P_{\pi,B_T\to\neg B_T}P_{\pi,\neg B_T\to B_T})U^{B_T}\notag \\
    +&P_{\pi,B_T\to\neg B_T}\begin{bmatrix}
    P_{\pi,\neg B_T\to B_A} & P_{\pi,\neg B_T\to\neg B_A} 
    \end{bmatrix}\mathbb{I}\notag \\
    +&P_{\pi,B_T\to\neg B_T} P_{\pi,\neg B_T\to \neg B_T}U^{\neg B_T} \notag \\
    \vdots\notag \\
    \overset{\textcircled{b}}{=}&(P_{\pi,B_T\to B_T}+P_{\pi,B_T\to\neg B_T}\sum_{k=0}^K{P_{\pi,\neg B_T\to\neg B_T}^k}P_{\pi,\neg B_T\to B_T})U^{B_T}\notag \\
    +&P_{\pi,B_T\to\neg B_T}\sum_{k=0}^K{P_{\pi,\neg B_T\to\neg B_T}^k}\begin{bmatrix}
    P_{\pi,\neg B_T\to B_A} & P_{\pi,\neg B_T\to\neg B_A} 
    \end{bmatrix}\mathbb{I}\notag \\
    +&P_{\pi,B_T\to\neg B_T} P_{\pi,\neg B_T\to\neg B_T}^{K+1}U^{\neg B_T} \notag 
\end{align}
where the equality $\textcircled{a}$ holds as we replace $U^{\neg B_T}$ in the last term by~\eqref{eqn:replace2}. 
Similarily, the equality $\textcircled{b}$ hold as we keep expanding $U^{\neg B_T}$ in the last term by~\eqref{eqn:replace2}. 
Since $P_{\pi,\neg B_T\to\neg B_T}$ only contains the transition probabilities between the transient states, for any $i,j\in \neg B_T$, $\lim_{K\to\infty}{(P_{\pi,\neg B_T\to\neg B_T}^K)_{ij}}=0$. Taking $K\to \infty$ in \eqref{eqn:replace2} and using the fact that $(I-P_{\pi,\neg B_T\to\neg B_T})^{-1} = \sum_{k=0}^\infty{P_{\pi,\neg B_T\to\neg B_T}^k}$, 
we have
\begin{align}\label{eqn:equality 1}
    &P_{\pi,B_T\to B_T}U^{B_T}+P_{\pi,B_T\to\neg B_T}U^{\neg B_T}\notag \\
        =&(P_{\pi,B_T\to B_T}+P_{\pi,B_T\to\neg B_T}(I-P_{\pi,\neg B_T\to\neg B_T})^{-1}P_{\pi,\neg B_T\to B_T})U^{B_T}\notag \\
    +&P_{\pi,B_T\to\neg B_T}(I-P_{\pi,\neg B_T\to\neg B_T})^{-1}\begin{bmatrix}
    P_{\pi,\neg B_T\to B_A} & P_{\pi,\neg B_T\to\neg B_A} 
    \end{bmatrix}\mathbb{I}.
\end{align}
Plugging \eqref{eqn:equality 1} into \eqref{eqn:Bellman tran}, we show the calculation of $U^{B_T}$ does not rely on $U^{\neg B}$,
\begin{align}
    U^{B_T} &=\gamma_B P_{\pi,B_T\to B_T}U^{B_T}+\gamma_BP_{\pi,B_T\to\neg B_T}U^{\neg B_T} +B_1\notag \\
    &=\gamma_B(P_{\pi,B_T\to B_T}+P_{\pi,B_T\to\neg B_T}(I-P_{\pi,\neg B_T\to\neg B_T})^{-1}P_{\pi,\neg B_T\to B_T})U^{B_T}\notag \\
    &+\gamma_BP_{\pi,B_T\to\neg B_T}(I-P_{\pi,\neg B_T\to\neg B_T})^{-1}\begin{bmatrix}
    P_{\pi,\neg B_T\to B_A} & P_{\pi,\neg B_T\to\neg B_A}
    \end{bmatrix}\mathbb{I} +B_1. 
\end{align}
Here, $P_{\pi}^{B_T}:=P_{\pi,B_T\rightarrow B_T} + P_{\pi,B_T\rightarrow \neg B_T}(I-P_{\pi,\neg B\to\neg B})^{-1}P_{\pi,\neg B_T\rightarrow B_T}$ where for any $i,j\in B_T$, $(P_{\pi}^{B_T})_{ij}$ is the probability of visiting $j$th state in $B$ without visiting any state in $B_T$ after leaving the $i$th state in $B_T$. 
As $P_{\pi}^{B_T}$ consists of only the transition probabilities between the transient states, 
$\lim_{k\to\infty}{({P_{\pi}^{B_T}}^k)_{ij}}=0$. Thus any eigenvalue $\lambda$ of ${P_{\pi}^{B_T}}$ satisfies $\vert \lambda\vert<1$.
Since $\gamma_B <1$, we are sure $(I-\gamma_B P_{\pi}^{B_T})$ is invertible and $U^{B_T}$ has a unique solution as, 
\begin{align}
    U^{B_T}&=(I-\gamma_B P_{\pi}^{B_T})^{-1} \gamma_BP_{\pi,B_T\to\neg B_T}(I-P_{\pi,\neg B\to\neg B})^{-1}\begin{bmatrix}
    P_{\pi,\neg B_T\to B_A} & P_{\pi,\neg B_T\to\neg B_A}
    \end{bmatrix}\mathbb{I} \notag \\  &+(I-\gamma_B P_{\pi}^{B_T})^{-1} B_1.
\end{align}
From \eqref{eqn:Bellman tran}, we have
\begin{align}
 U^{\neg B_T} &= P_{\pi,\neg B_T\rightarrow B_T}U^{B_T} + P_{\pi,\neg B_T\rightarrow \neg B_T} U^{\neg B_T} + B_2. 
\end{align} 
Using the fact $I- P_{\pi,\neg B_T\rightarrow \neg B_T}$ is invertible, we show $U^{\neg B_T}$ is uniquely determined by $U^{B_T}$ as 
\begin{align}
 U^{\neg B_T} = (I-P_{\pi,\neg B_T\rightarrow \neg B_T})^{-1}( P_{\pi,\neg B_T\rightarrow B_T}U^{B_T} + B_2),
\end{align} 
Thus, the equation \eqref{eqn:Bellman tran} has a unique solution. 
\fi

\end{document}